# Human-Like Autonomous Car-Following Model with Deep Reinforcement Learning

**Meixin Zhu[a,b], Xuesong Wang[a,b*], Yinhai Wang[c]**

[a]*Key Laboratory of Road and Traffic Engineering, Ministry of Education, Shanghai, 201804, China*

[b]*School of Transportation Engineering, Tongji University, Shanghai, 201804, China*

[c]*Department of Civil and Environmental Engineering, University of Washington, Seattle, WA 98195-2700, USA*

**ABSTRACT**

This study proposes a framework for human-like autonomous car-following planning based on deep reinforcement learning (deep RL). Historical driving data are fed into a simulation environment where an RL agent learns from trial and error interactions based on a reward function that signals how much the agent deviates from the empirical data. Through these interactions, an optimal policy, or car-following model that maps in a human-like way from speed, relative speed between a lead and following vehicle, and inter-vehicle spacing to acceleration of a following vehicle is finally obtained. The model can be continuously updated when more data are fed in. Two thousand car-following periods extracted from the 2015 Shanghai Naturalistic Driving Study were used to train the model and compare its performance with that of traditional and recent data-driven car-following models. As shown by this study's results, a deep deterministic policy gradient car-following model that uses disparity between simulated and observed speed as the reward function and considers a reaction delay of 1s, denoted as DDPGvRT, can reproduce human-like car-following behavior with higher accuracy than traditional and recent data-driven car-following models. Specifically, the DDPGvRT model has a spacing validation error of 18% and speed validation error of 5%, which are less than those of other models, including the intelligent driver model, models based on locally weighted regression, and conventional neural network-based models. Moreover, the DDPGvRT demonstrates good capability of generalization to various driving situations and can adapt to different drivers by continuously learning. This study demonstrates that reinforcement learning methodology can offer insight into driver behavior and can contribute to the development of human-like autonomous driving algorithms and traffic-flow models.

# 1 INTRODUCTION

Autonomous driving technology is capable of providing convenient and safe driving by avoiding crashes caused by driver errors (*1*). Considering, however, that we will likely be confronting a several-decade-long transition period when autonomous vehicles share the roadway with human-driven vehicles, it is critical to ensure that autonomous vehicles interact with human-driven vehicles safely. Because driving uniformity is a large factor in safety, a significant challenge of autonomous driving is to imitate, while staying within safety bounds, human driving styles, i.e., to achieve human-like driving (*2-4*).

Human-like driving will 1) provide passengers with comfortable riding, and confidence that the car can drive independently, and 2) enable surrounding drivers to better understand and predict autonomous vehicles' behavior so that they can interact with it naturally (*1*). To achieve human-like driving, it is useful to introduce a driver model that reproduces individual drivers' behaviors and trajectories.

Replicating driving trajectories is one of the primary objectives of modeling vehicles' longitudinal motion in traffic flow. Known as car-following models, these models are essential components of microscopic traffic simulation (*5*), and serve as theoretical references for autonomous car-following systems (*6*). Since the early investigation of car-following dynamics in 1953 (*7*), numerous car-following models have been built. Although these traditional car-following models have produced great achievements in microscopic traffic simulation, they have limitations in one or more of the following aspects when applied to autonomous car-following planning:

- First, limited accuracy. Most current car-following models are simplified, i.e., they contain only a small number of parameters (*8*). Simplification leads to suitable analytical properties and rapid simulations. However, it also renders the models limited in flexibility and accuracy because using few parameters can hardly model the inherently complex car-following process.

- Second, poor generalization capability. Car-following models calibrated with empirical data try to emulate drivers' output (e.g., speed and spacing) by finding the model parameters that maximize the likelihood of the actions taken in the calibration dataset. As such, the calibrated car-following model cannot be generalized to traffic scenarios and drivers that were not represented in the calibration dataset.

- Third, absence of adaptive updating. Most parameters of car-following models are fixed to reflect the average driver's characteristics (*9*). If car-following models with such pre-set parameters are used for autonomous vehicles, they can reflect neither the driving style of the actual drivers of the vehicles nor the contexts in which they drive.

To address the above limitations, a model is proposed for the planning of human-like autonomous car following that applies deep reinforcement learning (deep RL). Deep RL, which combines reinforcement learning algorithms with deep neural networks to create agents that can act intelligently in complex situations (*10*), has witnessed exciting breakthroughs such as deep *Q*-network (*11*) and AlphaGo (*12*). Deep RL has the promise to address traditional car-following models' limitations in that 1) deep neural networks are adept at general purpose function approximators (*13*) that can achieve higher accuracy in approximating the complicated relationship between stimulus and reaction during car-following; 2) reinforcement learning can achieve better generalization capability because the agent learns decision-making mechanisms from training data rather than parameter estimation through fitting the data (*11*); and 3) by continuously learning from historical driving data, deep RL can enable a car to move appropriately in accordance with the behavioral features of its regular drivers.

Figure 1 shows a schematic diagram of a human-like car-following framework based on deep RL. During a manual driving phase, data are collected and stored in the database as historical driving data. These data are then fed into a simulation environment where an RL agent learns from trial and error interactions by the environment including a reward function that signals how much the agent deviates from the empirical data. Through these interactions, an optimal policy, or car-following model is developed that maps in a human-like way from speed, relative speed between a lead and following vehicle, and inter-vehicle spacing to the acceleration of a following vehicle. The model, or policy, can be continuously updated when more data are fed in. This optimal policy will act as the executing policy in the autonomous driving phase.

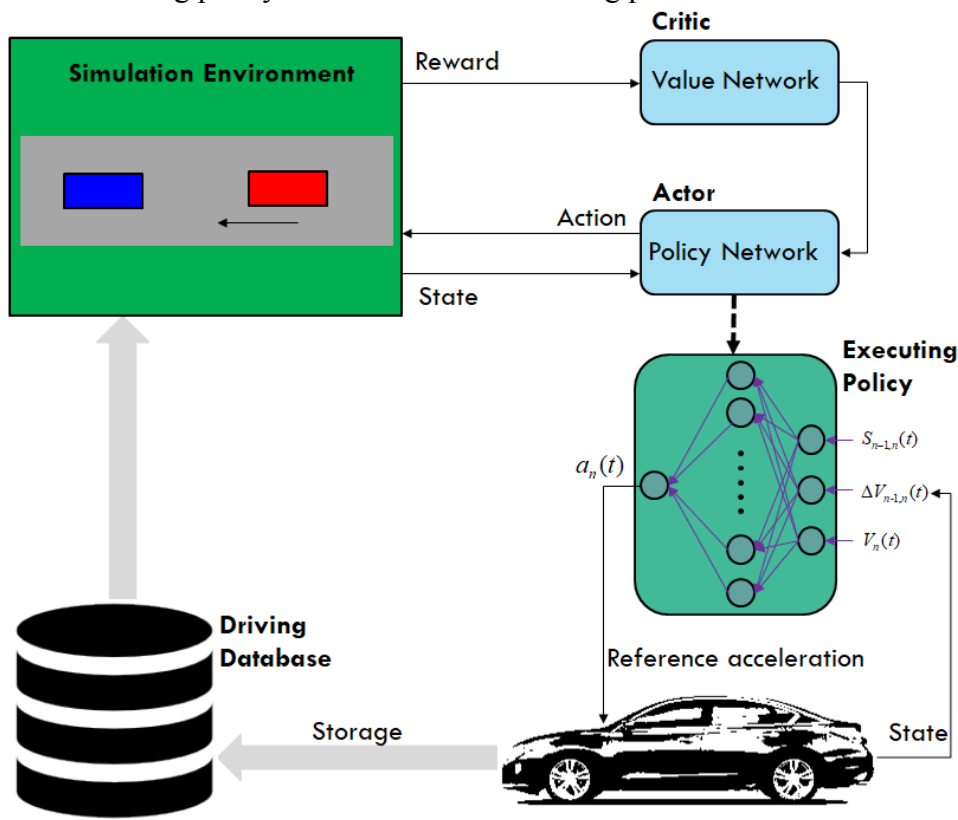

**FIGURE 1 Conceptual diagram of human-like car following by deep RL.**

To evaluate the proposed model and compare its performance with that of traditional car-following models, real-world driving data collected in the 2015 Shanghai Naturalistic Driving

Study (**Error! Reference source not found.**) were used to train and test the proposed deep RL-based model along with typical traditional car-following models and recent data-driven models. The models' performance in terms of trajectory-reproducing accuracy, generalization, and adaptivity were then compared.

This paper begins with a general background on car-following models and reinforcement learning. The data and the proposed deep RL-based car-following model are then specified, followed by training and performance evaluation of the model. The final section is devoted to discussion and conclusion.

## 2   BACKGROUND

### 2.1  Traditional Car-Following Models

A car-following model describes the motion of a following vehicle, on a single lane of roadway, with respect to the lead vehicle's actions. The first car-following models (*7, 15*) were proposed in the middle 1950s, and a number of models have been developed since then, for example, the Gaxis-Herman-Rothery (GHR) model (*16*), the intelligent driver model (IDM) (*17*), the optimal velocity model (*18*), and the models proposed by Helly (19), Gipps (*20*), and Wiedemann (*21*). For detailed review and historical development of the subject, consult Brackstone and McDonald (*5*), Olstam and Tapani (*22*), Panwai and Dia (*23*), and Saifuzzaman and Zheng (*24*); but in general, there exist four types of traditional car-following models: stimulus-based, safety-distance, psycho-physical and desired-measures models.

Stimulus-based models assume that the following vehicle determines its acceleration according to its relative distance and relative speed to the lead vehicle (*25*). The best known stimulus-based models are the General Motors models that have been proposed since the late 1950s, with the GHR model the most widely used.

The main idea of the safety-distance model is that the following driver selects a velocity and maintains a gap sufficient to stopping the vehicle safely should the lead vehicle make a sudden stop. The most widely used safety-distance model is the Gipps model (*20*).

Psycho-physical models assume that a driver varies his/her driving behavior according to traffic state by free driving, approaching the lead vehicle, stable following or emergency braking. Expressions combining relative speed and relative distance to the lead vehicle define the boundary conditions among different states (*21*). The Wiedemann model used in VISSIM® is a typical psycho-physical model.

The desired-measures models assume that a driver has some preferred situation that can be represented by certain measures (e.g., space headway and following speed), and that a driver continuously attempts to eliminate the difference between the actual situation and the preferred. One of the most widely used desired-measures models is the IDM (*26*). It considers both the desired speed and the desired following distance, as defined in Equation (*1*):

$$a_n\left(t\right) = a_{\max}^{(n)} \left[ 1 - \left( \frac{V_n\left(t\right)}{\tilde{V}_n\left(t\right)} \right)^{\beta} - \left( \frac{\tilde{S}_n\left(t\right)}{S_n\left(t\right)} \right)^2 \right] \qquad (1)$$

where $a_{\max}^{(n)}$ is the maximum acceleration/deceleration of the following vehicle $n$, $\tilde{V}_n$ is the desired speed, $S_n$ is the spacing between the two vehicles measured from the front edge of the following vehicle to the rear end of the lead vehicle ($S_n = \Delta X_n - L_n$ where $L_n$ is vehicle length), $\tilde{s}_n$ is the desired spacing, and $\beta$ is a parameter. The desired following distance in IDM is dependent on several factors: speed, speed difference ($\Delta V$), maximum acceleration ($a_{\max}^{(n)}$), comfortable deceleration ($a_{comf}^{(n)}$), minimum spacing at standstill ($S_{jam}^{(n)}$), and desired time headway ($\bar{T}_n$). Mathematically, the desired following distance can be calculated using Equation (2):

$$\tilde{S}_n\left(t\right) = S_{jam}^{(n)} + \max\left( 0, V_n\left(t\right)\tilde{T}_n\left(t\right) - \frac{V_n\left(t\right)\Delta V_n\left(t\right)}{2\sqrt{a_{\max}^{(n)} a_{comf}^{(n)}}} \right) \qquad (2)$$

This study compared the IDM with the proposed deep RL-based model because the IDM has been found to have the best performance among the traditional car-following models noted above (**Error! Reference source not found.**).

## 2.2 Data-Driven Car-Following Models

The recent availability of high-fidelity traffic data and consequential data-driven approaches provide an opportunity to model drivers' car-following behavior directly from mass field data. Data-driven methods are more flexible than traditional models, as they allow the incorporation of additional parameters that influence driving behavior, and thus lead to richer models. Previous data-driven car-following modeling studies can be divided into two main types: nonparametric regression and artificial neural network.

He et al. (*27*) proposed a simple nonparametric car-following model with k-nearest neighbor, which outputs the average of the most similar cases, i.e., the most likely driving behavior under the current circumstance. Papathanasopoulou and Antoniou (*28*) developed a nonparametric data-driven car-following model based on locally weighted regression, the Loess model.

Modeling car-following behavior with artificial neural networks dates back to 2003 when Jia et al. (*29*) introduced a four-layer neural network (including one input layer, two hidden layers, and one output layer). This neural network takes relative speed, desired speed, follower speed and gap distance as inputs, using them to predict follower acceleration. Chong et al. (*30*) illustrated that it is possible to predict acceleration accurately using neural networks with only one hidden layer, and Khodayari et al. (*31*) improved the work of Chong et al. (*30*) by considering instantaneous reaction time (RT) delay. In contrast to these conventional neural network-based models, recent studies have begun to model car-following behavior with recurrent neural networks (RNN) (*32*), and with more than one hidden layer, i.e., deep neural networks (*8*).

The current study is the first to use a deep reinforcement learning to model car-following behavior. Of the other recent data-driven car-following models, the RNN, nonparametric Loess, and conventional neural network-based (designated hereafter as NNa) models were selected for comparison.

## 2.3 Reinforcement Learning

By setting up simulations in which an agent interacts with an environment over time, reinforcement learning (RL) aims to address problems related to sequential decision making. At time step $t$, an RL agent observes a state $s_t$ and chooses an action $a_t$ from an action space $A$, according to a policy $\pi(a_t|s_t)$ that maps from states to actions. The agent gets a reward $r_t$ for the action choice and moves to the next state $s_{t+1}$, following the environmental dynamic, or model. When the agent reaches a terminal state, this process stops and then restarts. The task is to maximize the expectation of a discounted, accumulated reward $R_t = \sum_{k=0}^{\infty} \gamma^k r_{t+k}$ from each state, where $\gamma \in (0,1]$ represents the discount factor (*33*).

Specifically, in the modeling of car-following behavior, the state at a certain time step $t$ is described by the following key parameters: the speed of a following vehicle $V_n(t)$, inter-vehicle spacing $S_{n-1,n}(t)$, and relative speed between a lead and following vehicle $\Delta V_{n-1,n}(t)$. The action is the longitudinal acceleration of the following vehicle $a_n(t)$, confined between -3 m/s$^2$ and 3 m/s$^2$ based on the acceleration of all observed car-following events. The reward value is given based on deviations between predicted and observed car-following trajectories. With state and action at time step $t$, a kinematic point-mass model was used for state updating:

$$V_n(t+1) = V_n(t) + a_n(t) \cdot \Delta T$$

$$\Delta V_{n-1,n}(t+1) = V_{n-1}(t+1) - V_n(t+1) \tag{3}$$

$$S_{n-1,n}(t+1) = S_{n-1,n}(t) + \frac{\Delta V_{n-1,n}(t) + \Delta V_{n-1,n}(t+1)}{2} \cdot \Delta T$$

where $\Delta T$ is the simulation time interval, set as 0.1s in this study, and $V_{n-1}$ is the velocity of lead vehicle (LV), which was externally inputted.

### 2.3.1 Value-Based Reinforcement Learning

A value function predicts the expected long-term reward, measuring the quality of each state, or state-action pair. The action value $Q^\pi(s,a) = E[R_t \mid s_t = s, a_t = a]$ stands for the expected reward for choosing action $a$ in state $s$ and then following policy $\pi$; the value represents the quality of a certain action $a$ in a given state $s$. Value-based RL methods aim to learn the action value function from experience: one example of such an algorithm is $Q$-learning. Beginning with a random $Q$-function, the agent updates the $Q$-function based on the Bellman equation (*34*):

$$Q(s,a) = E[r + \gamma \max_{a'} Q(s',a')] \qquad (4)$$

This equation is based on the following intuition: the immediate reward $r$ plus maximum future reward for the next state $s'$ is the maximum future reward for state $s$ and action $a$. The agent chooses the action with the highest $Q(s,a)$ to get maximum expected future returns.

### 2.3.2    Policy-Based Reinforcement Learning

Based on function approximation, policy-based methods calculate the gradient ascent on $E[R_t]$ directly and update the parameters $\theta$ of policy $\pi(a \mid s;\theta)$ in the direction of $\nabla_\theta \log \pi(a_t \mid s_t;\theta)R_t$.

To update the gradients with lower variance and faster speed, an actor-critic method is usually used, in which the algorithm is comprised of two learning agents, the actor (policy) and the critic (value function). The actor functions to generate an action $a$, given the current state $s$, and the critic is in charge of evaluating the actor's current policy. The critic updates and approximates the value function using samples. The updated value function is then used to improve the actor's policy (*35*).

## 2.4  Deep Reinforcement Learning

When a value function $Q(s;\theta)$, policy $\pi(a \mid s;\theta)$, or an environmental model is represented by neural networks, the reinforcement learning algorithms are referred as deep reinforcement learning, where $\theta$ indicates weights in a deep neural network. For example, in car-following modeling , a neural network-based actor takes a state $s_t = (v_n(t), \Delta v_{n-1,n}(t), \Delta S_{n-1,n}(t))$ as input, and outputs a continuous action: the following vehicle's acceleration $a_n(t)$.

### 2.4.1    Deep Q-Network

Deep $Q$-Learning (*11*) is a typical deep RL algorithm that performs in discrete action spaces. It uses a deep network to predict the value function of each discrete action and chooses the action with maximum value output. Deep $Q$ networks (*DQN*) work well when there are only a few possible actions, but can fail in continuous action spaces. To resolve such difficulties, the deep deterministic policy gradient (DDPG) algorithm was proposed by Lillicrap et al. (*36*). The DDPG builds on *DQN* but can be applied to continuous-control problems.

### 2.4.2    Deep Deterministic Policy Gradient

Like *DQN*, DDPG uses deep neural network function approximators. Unlike *DQN*, however, DDPG uses two separate actor and critic networks (*36*). The critic network with weights $\theta^Q$ approximates the action-value function $Q(s,a \mid \theta^Q)$. The actor network with weights $\theta^\mu$ explicitly represents the agent's current policy $\mu(s|\theta^\mu)$ for the $Q$-function, which maps from a state of the environment $s$ to an action $a$. To enable stable and robust learning, DDPG deploys experience replay and target network, as in *DQN*:

**(1) Experience replay**

A replay buffer is applied to address the issue that training samples in RL are not independently and identically distributed because they are generated by sequential explorations in an environment. The replay buffer is a cache $D$ of finite size. By sampling transitions ($s_t$, $a_t$, $r_t$, $s_{t+1}$) based on the model's exploration policy (see 4.5 below), historical data are stored in the replay buffer. The oldest samples are replaced with new when the replay buffer is full. The actor and critic are updated with a minibatch sampled from the buffer.

**(2) Target network**

To address the divergence of direct implementation of $Q$ learning with neural networks, separate target networks responsible for calculating the target values are used. In the DDPG algorithm, two target networks, $Q'(s,a|\theta^{Q'})$ and $\mu'(s|\theta^{\mu'})$, are created for the main critic and actor networks, respectively. They are identical in shape to the main networks but have different network weights $\theta'$. The target networks are updated in a way that slowly tracks the learned networks: $\theta' = \tau\theta + (1-\tau)\theta'$ with $\tau \square 1$. By constraining target values with slow updating speeds, stable learning can be achieved.

This study proposed to model car-following behavior with deep reinforcement learning because it may address traditional car-following models' limitations in two aspects: 1) deep neural networks can achieve higher accuracy in approximating the relationship between stimulus and reaction during car-following; and 2) reinforcement learning can achieve better generalization capability by learning decision-making mechanisms from training data rather than parameter estimation through data fitting. The proposed model was expected to have better performance than traditional car-following models in terms of trajectory-reproducing accuracy and generalization capability.

## 3   DATA PREPARATION

### 3.1  Shanghai Naturalistic Driving Study

Driving data collected in the Shanghai Naturalistic Driving Study (SH-NDS) (***Error! Reference source not found.***) were used in this study. Jointly conducted by Tongji University, General Motors, and the Virginia Tech Transportation Institute, the SH-NDS used five vehicles to collect real-world driving data from December 2012 to December 2015. Data from 60 Chinese drivers across a total mileage of 161,055 km were collected in the SH-NDS study.

## 3.2 Data Acquisition System

The SH-NDS used Strategic Highway Research Program 2 (SHRP2) NextGen data acquisition systems (DAS) to collect the following data items at frequencies ranging from 10 to 50 Hz: position data from the vehicles' global positioning systems (GPS), longitudinal and lateral acceleration from their accelerators, data such as steering and throttle angle from a vehicle controller area network (CAN, a robust vehicle standard network designed to allow microcontrollers and devices to communicate with each other), relative distances and speeds to surrounding vehicles from their radar systems, and the vehicles' four synchronized camera views. Figure 2 presents the four views, which include a face view for the driver's facial expressions, front and rear views for the roadway ahead and behind the vehicle, and a hand view for the driver's hand maneuvers.

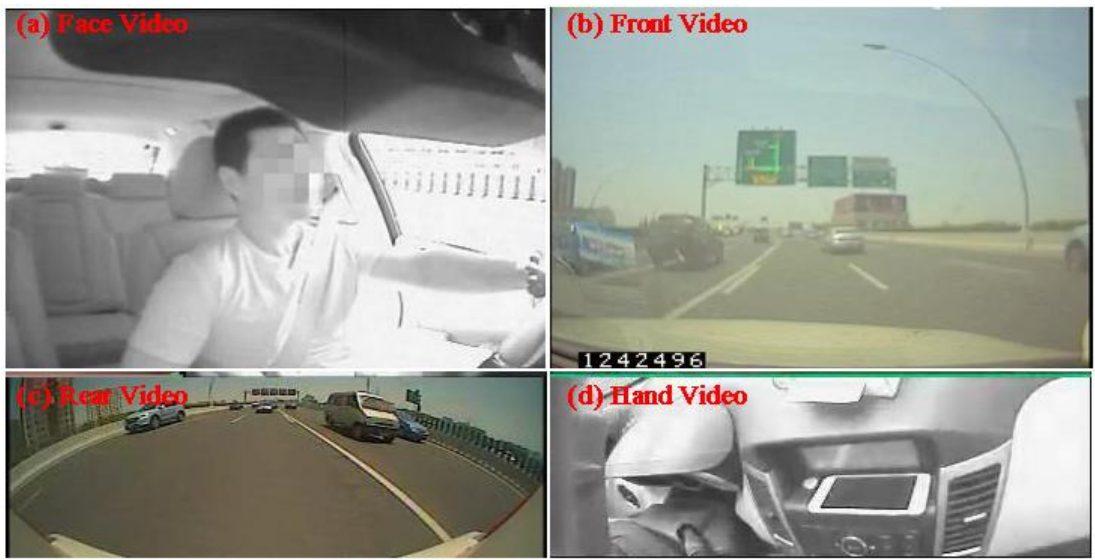

**FIGURE 2 SH-NDS four camera views.**

## 3.3 Car-Following Periods Extraction

Car-following periods were automatically extracted by applying a car-following filter, described in a previous study by Zhu et al. (***Error! Reference source not found.***):

- Radar target's identification number remained constant: indicating the subject vehicle was preceded by the same lead vehicle (LV);
- Longitudinal distance <120 m: eliminating free-flow traffic conditions;
- Lateral distance <2.5 m: ensuring that the two vehicles were in the same lane; and
- Car-following period duration >15 s: guaranteeing that the car-following period contained enough data to be analyzed.

 The results of the automatic extraction process were confirmed with video viewed by an analyst to ensure the validity of the car-following periods selected for analysis. Because this study focused on drivers' car-following behavior on roadways with limited access, the roadway type was identified during the validation phase so only data from car-following periods on limited access expressways and freeways were used.

Considering that training DDPG and the comparison models requires substantial time and computing resources, study data was further limited to the car-following periods of 20 drivers randomly selected from the original 60 SH-NDS drivers. For each of the 20 drivers, 100 periods were randomly selected; 70 of those periods were randomly selected for calibration and 30 for validation. The total 2,000 selected car-following periods represented 827 minutes of driving.

As evidenced by the variation in following gap, speed, relative speed, time gap, and acceleration, the 20 investigated drivers displayed substantially diverse driving behavior. Therefore, by selecting by mean and standard deviation of speed, gap, and relative speed feature vectors, the drivers were clustered into two driving styles, aggressive and conservative, using the *k*-means algorithm. Figure 3 presents the detailed car-following trajectories in (a) speed versus gap and (b) speed versus time gap. As can be seen, aggressive drivers (in blue) maintained shorter spacing and time gaps than conservative drivers (in red).

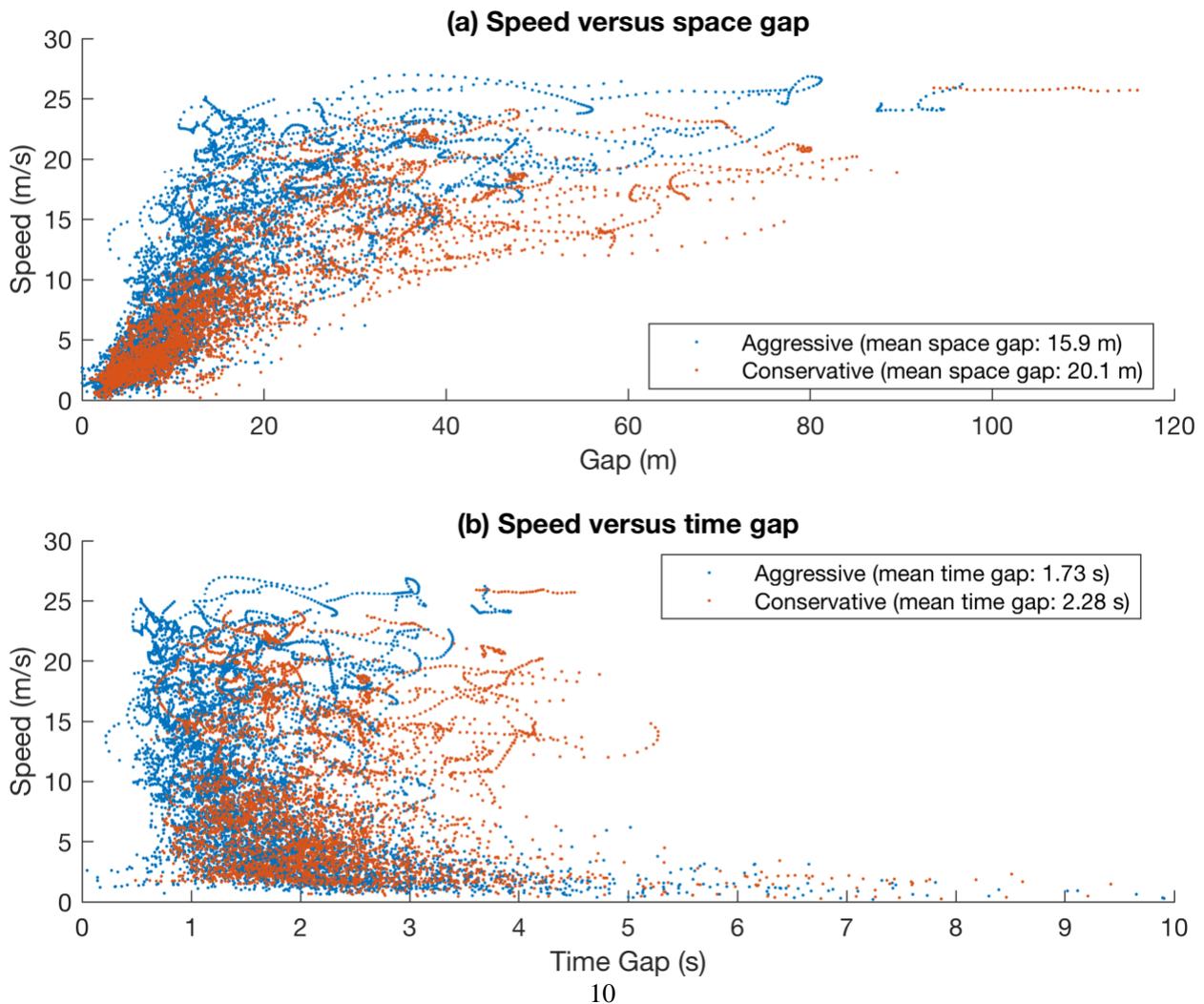

**FIGURE 3 Car-following trajectories for aggressive and conservative drivers.**

The cluster display of driving behavior diversity was further confirmed by descriptive statistics of the drivers' car-following events. As shown in Table 1, aggressive drivers had shorter mean space and time gap than conservative drivers with similar mean following speeds.

**TABLE 1 Descriptive Statistics for Different Driving Styles in Car-Following Events**

| Type | Space Gap (m) | | | Speed (m/s) | | | Absolute Relative Speed (m/s) | | |
|------|------|-----|-----|------|-----|-----|------|-----|-----|
| | Mean | Min | Max | Mean | Min | Max | Mean | Min | Max |
| Aggressive | 15.90 | 0.16 | 96.77 | 10.50 | 0 | 27.00 | 0.76 | -4.90 | 4.20 |
| Conservative | 20.10 | 0.77 | 115.95 | 9.20 | 0 | 25.97 | 0.81 | -4.82 | 4.90 |
| All | 17.53 | 0.16 | 115.95 | 9.99 | 0 | 27.00 | 0.78 | -4.90 | 4.90 |

| Type | Time Gap (s) | | | Absolute Acceleration (m/s²) | | | Time Fraction of Stopped Traffic | | |
|------|------|-----|-----|------|-----|-----|------|-----|-----|
| | Mean | Min | Max | Mean | Min | Max | | | |
| Aggressive | 1.73 | 0.63 | 22.39 | 0.35 | -3.65 | 2.37 | 1.39% | | |
| Conservative | 2.28 | 0.43 | 63.69 | 0.31 | -2.94 | 2.03 | 1.38% | | |
| All | 1.94 | 0.43 | 63.69 | 0.34 | -3.65 | 2.37 | 1.38% | | |

# 4    APPROACH TO THE PROPOSED MODEL

Because the acceleration of a vehicle is continuous, the deep RL method of deep deterministic policy gradient (DDPG) (*36*), which performs well in continuous action space, was used. In this section, the DDPG approach to modeling car-following behavior is explained.

## 4.1  Simulation Setup

A simple car-following simulation environment was implemented to enable the RL agent to interact with the environment through a sequence of states, actions, and rewards. Data from the lead vehicle served as externally controlled input. The following vehicle's speed, spacing and velocity differences were initialized with the empirical SH-NDS data: $V_n(t=0) = V_n^{data}(t=0)$ and $S_{n-1,n}(t=0) = S_{n-1,n}^{data}(t=0)$. After the acceleration $a_n(t)$ was computed by the RL agent, future states of the following vehicle were generated iteratively, based on the state-updating rules defined in Equation (*3*). The simulated spacings, or gaps, were compared to the empirical spacing from the SH-NDS data to calculate reward values and simulation errors. The state was re-initialized with the empirical dataset when the simulated car-following event terminated at its maximum time step.

## 4.2  Evaluation Metric and Reward Function

As suggested in Punzo and Montanino (*37*), the empirical and simulated inter-vehicle spacings and speeds were compared in this study to evaluate each car-following model's performance. Specifically, the root mean square percentage error (RMSPE) of spacing and speed, or velocity, was adopted as the evaluation metric:

$$\text{RMSPE of spacing} = \sqrt{\frac{\sum_{i=1}^{N}(S_i^{sim} - S_i^{obs})^2}{\sum_{i=1}^{N}(S_i^{obs})^2}} \qquad (5)$$

$$\text{RMSPE of speed} = \sqrt{\frac{\sum_{i=1}^{N}(V_i^{sim} - V_i^{obs})^2}{\sum_{i=1}^{N}(V_i^{obs})^2}} \qquad (6)$$

where $i$ denotes observation, $S_i^{sim}$ and $V_i^{sim}$ are the $i$th modeled spacing and speed, $S_i^{obs}$ and $V_i^{obs}$ are the $i$th observed spacing and speed, and $N$ is the number of observations.

In RL, the reward function $r(s, a)$ provides a scalar value indicating the desirability of a particular state transition from the initial state $s$ to a successor state $s'$ resulting from performing action $a$. This study used the reward function to facilitate human-like driving by minimizing the disparity between the values of simulated and observed behavior. As results could differ depending on the type of reward function, both spacing and speed (velocity) were tested:

$$r_t = \log\left(\left|\frac{S_{n-1,n}(t) - S_{n-1,n}^{obs}(t)}{S_{n-1,n}^{obs}(t)}\right|\right) \qquad (7)$$

$$r_t = \log\left(\left|\frac{V_n(t) - V_n^{obs}(t)}{V_n^{obs}(t)}\right|\right) \qquad (8)$$

where $S_{n-1,n}(t)$ and $V_n(t)$ are the simulated spacing and speed, respectively, in the RL environment at time step $t$, and $S_t^{obs}$ and $V_n^{obs}(t)$ are the observed spacing and speed in the empirical data set at time step $t$. The DDPG model adopting spacing disparity as the reward function is represented by DDPGs, and the DDPGv model represents the reward function of speed disparity.

### 4.3 Network Architecture

Two separate neural networks were used to represent the actor and critic. At time step $t$, the actor network takes a state $s_t = (v_n(t), \Delta v_{n-1,n}(t), \Delta S_{n-1,n}(t))$ as input, and outputs a continuous action: the following vehicle's acceleration $a_n(t)$. The critic network takes a state $s_t$ and an action $a_t$ as input, and outputs a scalar $Q$-value $Q(s_t, a_t)$.

As presented in Figure 4, the actor and critic networks each have three layers: an input layer, an output layer, and a hidden layer containing 30 neurons between the input and output layers. This study experimented with neural networks deeper than one hidden layer, but the results showed that they were unnecessary for our purposes, which demanded only three or four input variables.

In the hidden layers, each neuron has a rectified linear unit (RLU) activation function that transforms its input to its output signal. The RLU function computes as $f(x) = \max(0, x)$ and has been shown to accelerate the convergence of network parameter optimization significantly (*38*). The final output layer of the actor network used a tanh activation function, which maps a real-valued number to the range [-1, 1], and thus can be used to bound the output actions.

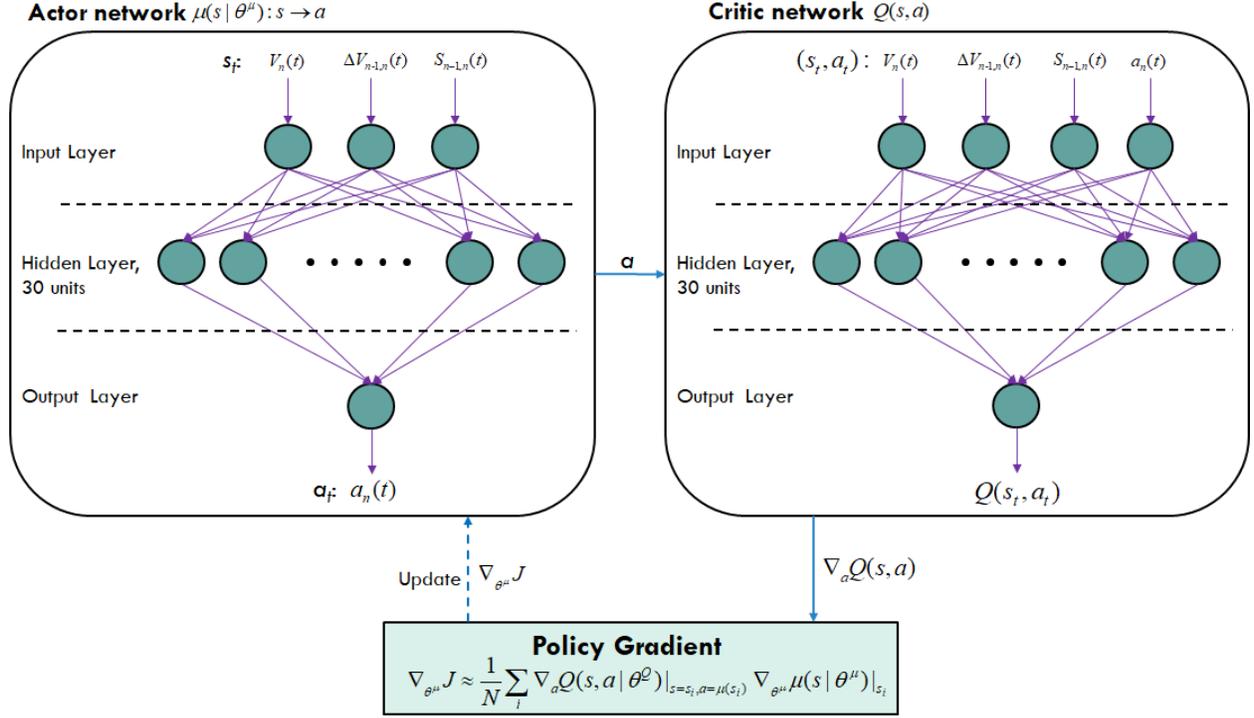

**FIGURE 4 Architecture of the actor and critic networks.**

To test the impact of driver reaction time (RT) on following speed and spacing, this study also proposed models, designated as DDPGvRT and DDPGsRT respectively, that take time-series data as input. A reaction time of 1s was adopted in accordance with Zhou et al. (*32*). For these models that considered reaction time, the number of hidden layer neurons for actor and critic networks increased to 100 to account for the greater number (30 or 40) of inputs.

## 4.4 Network Update and Hyperparameters

At each learning step, the weight coefficients of the critic network were updated using the stochastic gradient descent algorithm in order to minimize the loss function $L = \frac{1}{N} \sum_i (y_i - Q(s_i, a_i \mid \theta^Q))^2$. The adaptive learning rate trick Adam (*39*) was also applied. As shown in Figure 4, the actor network parameters were updated as follows: the acceleration outputted by the actor was passed forward to the critic; next, the gradients $\nabla_a Q(s, a)$ indicated how the action should be updated to increase the $Q$-Value; these gradients were then passed to the actor, updating the actor's parameters $\theta^\mu$ according to the gradients

$$\nabla_{\theta^\mu} J \approx \frac{1}{N} \sum_i \nabla_a Q(s, a \mid \theta^Q) \mid_{s=s_i, a=\mu(s_i)} \nabla_{\theta^\mu} \mu(s \mid \theta^\mu) \mid_{s_i}.$$

The adopted hyperparameters, parameters whose values are set prior to the beginning of the learning process, are shown in Table 2. The values of these hyperparameters were determined by testing a randomly sampled training dataset, in accordance with Lillicrap et al. (*36*).

**TABLE 2 Hyperparameters and Corresponding Descriptions**

| Hyperparameter | Value | Description |
|---|---|---|
| Learning rate | 0.0005 | Learning rate used by Adam |
| Discount factor | 0.9 | Q-learning discount factor gamma |
| Minibatch size | 256 | Number of training cases used by stochastic gradient descent update |
| Replay start size | 7000 | Number of time steps used for running random policy before learning starts; resulting experiences are stored in replay memory |
| Replay memory size | 10000 | Number of training cases in replay memory |
| Soft target update $\tau$ | 0.01 | Update rate of target networks |

## 4.5 Exploration Noise

The exploration-exploitation trade-off is a fundamental dilemma in reinforcement learning: exploitation refers to making the best decision using past knowledge while exploration refers to the pursuit and acquisition of new knowledge by taking exploratory actions. The need for exploitation is obvious, but because the best long-term policy may involve short-term sacrifices, exploitation only is not enough. Exploration is essential for gathering enough information to make the best overall decisions. The current study therefore constructed an exploration policy by adding a noise process to the original actor policy. The Ornstein-Uhlenbeck process (*39*) with $\theta = 0.15$ and $\sigma = 0.2$ was used, as suggested by Lillicrap et al. (*36*). The process produces temporally correlated values centered around zero, enabling good exploration in physical environments.

## 4.6 Full Algorithm

The full DDPG algorithm for car-following modeling is Algorithm 1 below. DDPG starts by initializing the replay buffer as well as its actor, critic, and corresponding target networks. For each episode step, the follower acceleration is calculated according to the actor policy. Afterwards, the reward $r_t$ and new state $s_{t+1}$ are observed and the information is stored in the replay memory $D$.

During training, minibatches are sampled from replay memory. The critic is then updated with the same loss function as in *DQN*. In the next step, the actor is updated by performing an ascent step on the sampled policy gradient. Finally, the target networks with weights $\theta^{Q'}$ and $\theta^{\mu'}$ are slowly moved in the direction of the updated weights of the actor and critic networks.

**Algorithm 1** DDPG: Deep deterministic policy gradient for car-following modeling (*36*)

Randomly initialize critic $Q(s, a \mid \theta^Q)$ and actor $\mu(s \mid \theta^\mu)$ networks with weights $\theta^Q$ and $\theta^\mu$.

Initialize target network $Q'(s, a \mid \theta^{Q'})$ and $\mu'(s \mid \theta^{\mu'})$ with weights $\theta^{Q'} \leftarrow \theta^Q, \theta^{\mu'} \leftarrow \theta^\mu$

Set up empty replay buffer $D$

**for** episode = 1 to M **do**

    Begin with a random process $\mathbb{N}$ for action exploration

    Observe initial car-following state $s_1$: initial gap, follower speed, and relative speed

    **for** t = 1 to T **do**

        Calculate reward $r_t$ based on disparity between values of simulated and observed behavior

        Choose follower acceleration $a_t = \mu(s_t \mid \theta^\mu) + \mathbb{N}_t$ based on current actor network and exploration noise $\mathbb{N}_t$

        Implement acceleration $a_t$ and transfer to new state $s_{t+1}$ based on kinematic point-mass model

        Save transition $(s_t, a_t, r_t, s_{t+1})$ into replay buffer $D$

        Sample random minibatch of $N$ transitions $(s_i, a_i, r_i, s_{i+1})$ from $D$

        Set $y_i = r_i + \gamma Q'(s_{i+1}, \mu'(s_{i+1} \mid \theta^{\mu'}) \mid \theta^{Q'})$

        Update critic through minimizing loss: $L = \dfrac{1}{N} \sum_i (y_i - Q(s_i, a_i \mid \theta^Q))^2$

        Update actor policy using sampled policy gradient:

$$\nabla_{\theta^\mu} J \approx \frac{1}{N} \sum_i \nabla_a Q(s, a \mid \theta^Q)|_{s=s_i, a=\mu(s_i)} \nabla_{\theta^\mu} \mu(s \mid \theta^\mu)|_{s_i}$$

        Update target networks:

$$\theta^{Q'} = \tau\theta^Q + (1-\tau)\theta^{Q'}$$
$$\theta^{\mu'} = \tau\theta^\mu + (1-\tau)\theta^{\mu'}$$

    **end for**

**end for**

## 5 TRAINING AND TEST

### 5.1 Test Procedure

The car-following models were calibrated, or trained, at an individual-driver level, i.e., the calibration process was repeated for each driver, who each had his/her own set of parameters. As shown in Figure 5, each calibration and validation proceeded as follows:

1) Seventy of the total 100 car-following periods (70%) were randomly selected for the calibration dataset.

2) The car-following models were trained based on the calibration dataset, outputting model parameters.

3) Two levels of validation—intra-driver and inter-driver—were conducted for the calibrated model parameters.

    A. Intra-driver validation aimed to assess how the calibrated models would perform when

applied to car-following periods that were not used for calibration but belonged to the currently investigated, or calibrated, driver. That is, the remaining 30 car-following periods of the calibrated driver were used for intra-driver validation. Intra-driver validation error was used to measure the trajectory-reproducing accuracy of each car-following model.

B. Inter-driver validation aimed to assess how the calibrated models would perform when applied to other drivers. For each calibrated driver, inter-driver validation was repeated for each of the remaining 19 validating drivers, with all 100 car-following periods of each validating driver used in each repetition. Inter-driver validation error was used to measure the generalization capability for all car-following models.

4) For the DDPG car-following models, a model originally trained with data from one driver, e.g., Driver A, was then retrained with data from another driver, Driver B. The error reduction induced by the retraining was used to show how the DDPG model can adapt to different drivers.

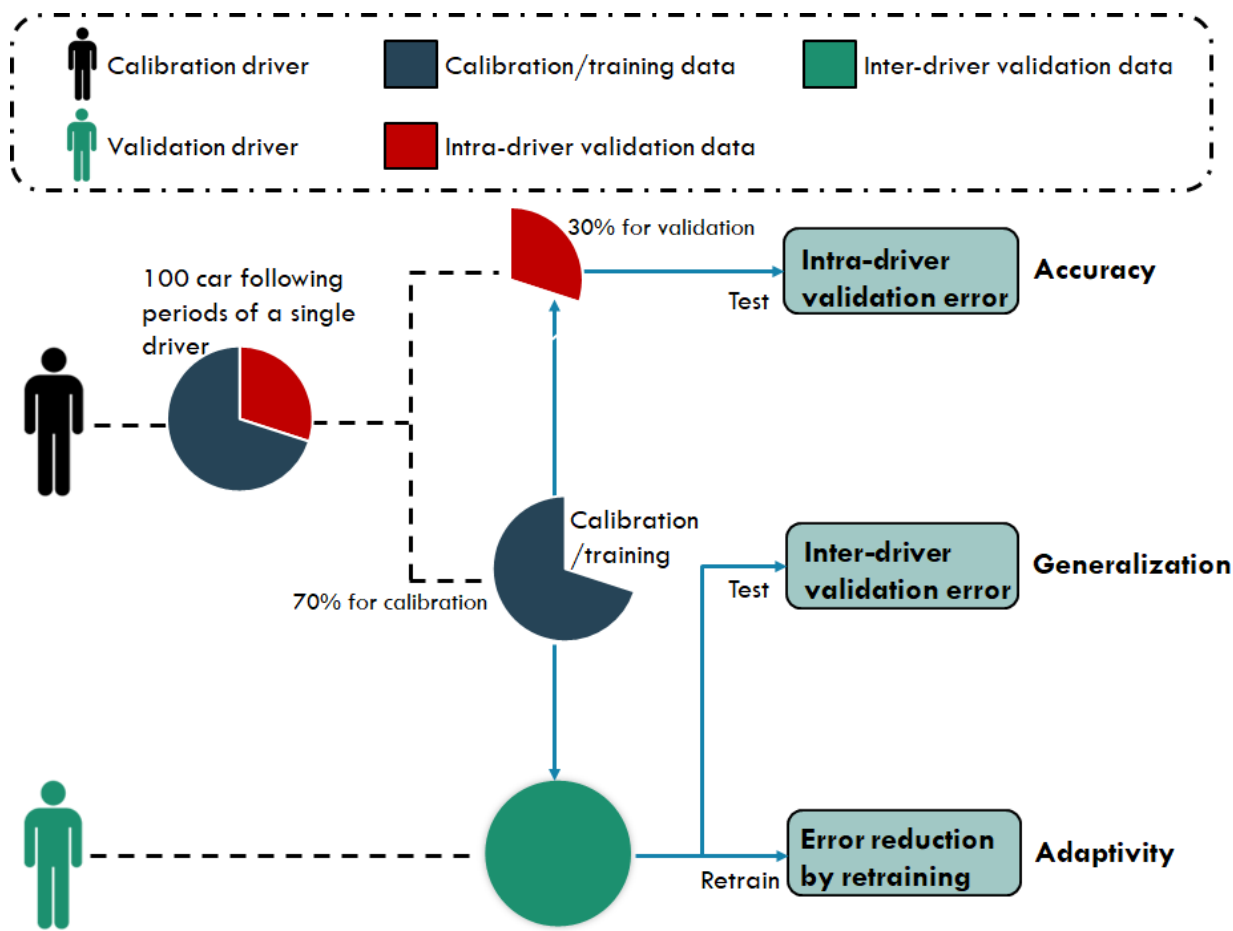

**FIGURE 5 Schematic representation of the calibration and validation process for a single driver.**

## 5.2 Models Investigated

As shown in Table 3, the current study compared the performance of 7 car-following models:

DDPGs, DDPGv and DDPGvRT are the proposed models, which differ in reward function type (spacing or velocity) and reaction time (RT) delay; IDM is representative of traditional car-following models; and NNa, which has a network architecture the same as that of DDPGv's actor network, RNN, and Loess are representative of recent data-driven models. The performance of DDPGsRT and NNaRT are not reported in the table because they performed worse than the corresponding models that did not consider reaction time delay, DDPGs and NNa. Detailed descriptions of the Loess, NNa, and RNN car-following models are presented in the appendix.

**TABLE 3 Description of the Investigated Car-Following Models**

| Model | Description |
|-------|-------------|
| DDPGs | Deep deterministic policy gradient (DDPG) model with spacing deviation as reward function, without reaction delay |
| DDPGv | DDPG model with speed (velocity) deviation as reward function, without reaction delay |
| DDPGvRT | DDPG model with speed deviation as reward function, considering a reaction delay of 1s |
| IDM | Intelligent driver model, representing traditional car-following models |
| RNN (*32*) | Car-following model based on recurrent neural networks |
| Loess (*28*) | Car-following model based on locally weighted regression, representing data-driven nonparametric models |
| NNa | Car-following model that predicts follower acceleration, representing data-driven conventional neural network-based models |

### 5.3 Calibrating the IDM

Car-following models are calibrated to find model parameters that minimize the disparity between simulated and observed values (*41*). In this study, the root mean square percentage error (RMSPE) of spacing defined in Equation (*5*) was used as the error measure, and a genetic algorithm (GA) was implemented to find the optimum parameter values.

The Genetic Algorithm Toolbox in MATLAB® was used to calibrate the IDM in this study. The relevant GA parameters were set as follows: maximum number of generations at 300, number of stall generations at 100, and population size at 300. Because the GA uses a stochastic process, it outputs slightly different optimal parameters in different optimization runs. Twelve runs were conducted for each driver, and the set of parameters with the minimum error (RMSPE) was chosen. For detailed information about the calibration, readers can refer to Zhu et al. (***Error! Reference source not found.***).

### 5.4 Training the DDPG Car-Following Models

As noted above, 70 car-following periods were randomly selected to train the DDPG car-following models for each driver, and the remaining 30 periods were used as test, or validation, data. For each training episode, the 70 car-following periods were simulated sequentially by the RL agent.

The state was initialized according to the empirical data whenever a new car-following period was to be simulated, and the spacing RMSPE for both training and test data were calculated during training. The training was repeated for 60 episodes for each driver, and the RL agent that generated the smallest sum of training and test errors was selected. Figure 6 shows the reward and error curves of the DDPGs model for a randomly selected driver. As can be seen, the performance of the DDPGs model starts to converge when the training episode reaches 20.

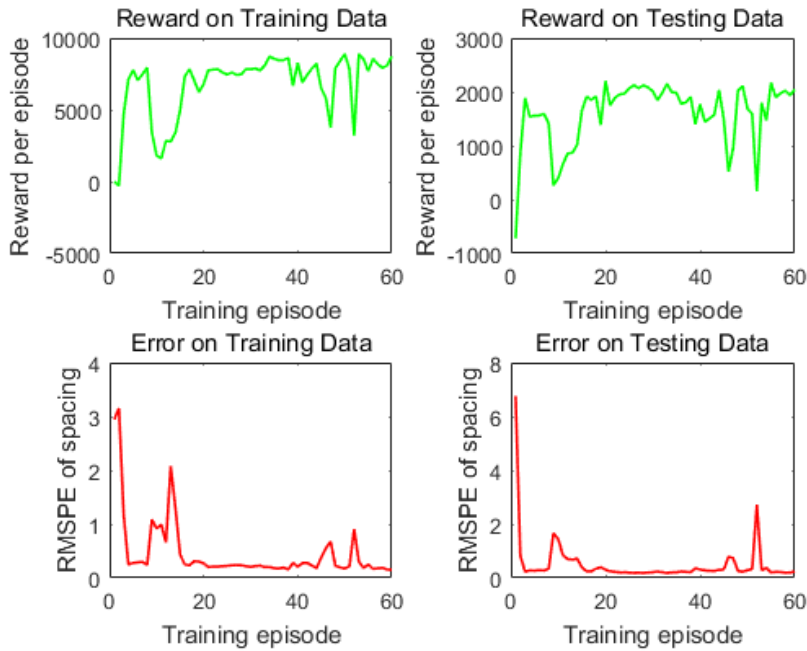

**FIGURE 6 Reward and error curves of DDPGs model for a random driver.**

# 6   RESULTS

## 6.1   Trajectory-Reproducing Accuracy

To examine the driving trajectory-reproducing accuracy of the proposed DDPG car-following models, we compared the models' RMSPE of spacing and speed in the intra-driver validation datasets.

Figure 7 presents the mean value and standard deviation of the intra-driver validation errors for each of the seven models. The DDPGvRT model outperformed all the investigated models, as evidenced by its having the lowest mean and standard deviation on spacing and speed.

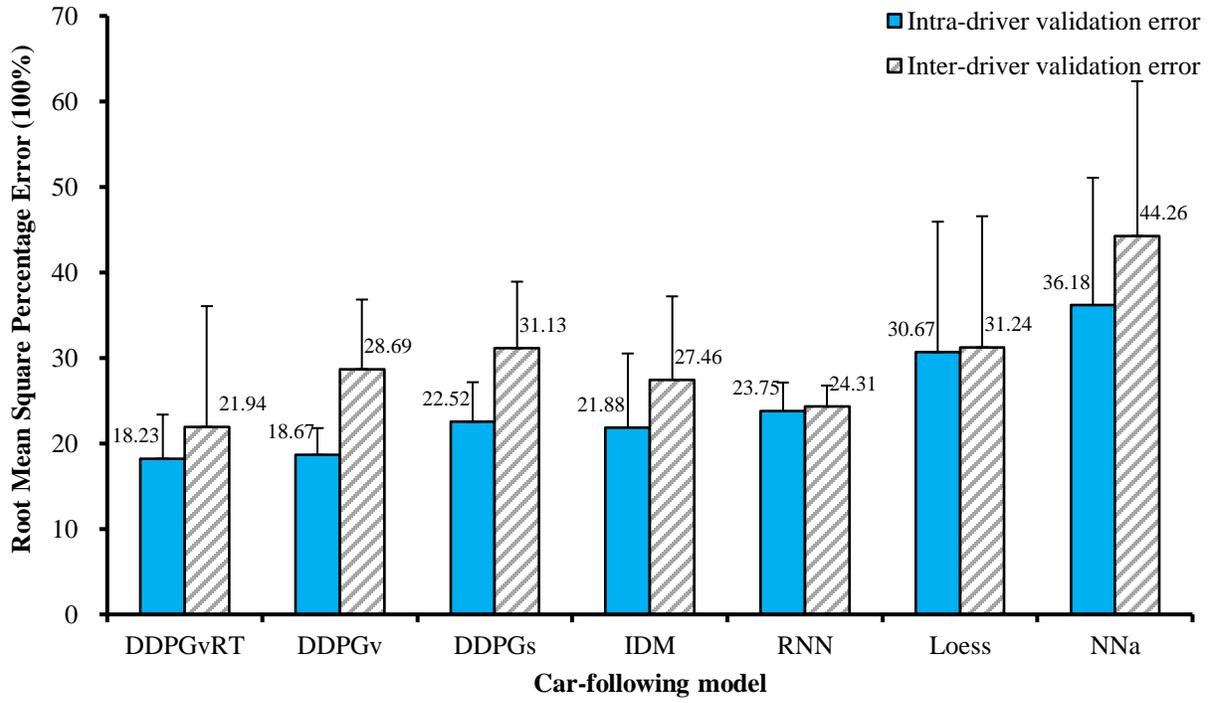

**(a) Spacing**

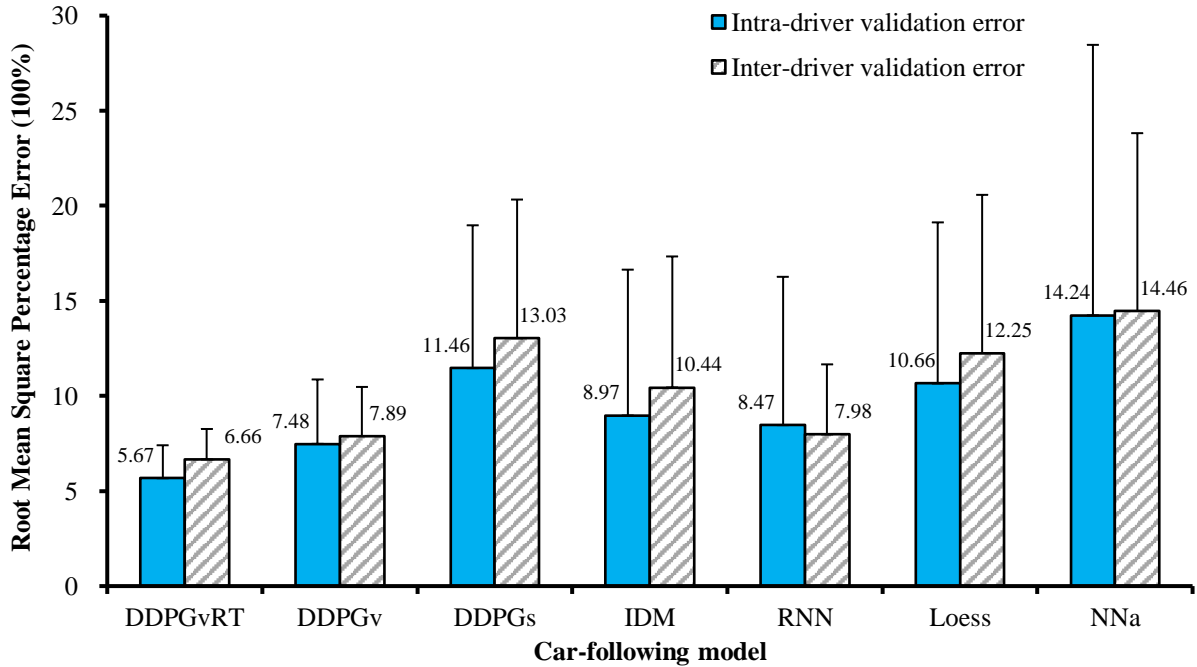

**(b) Speed**

**FIGURE 7 Mean and standard deviation (verticals above the bars) of intra- and inter-driver validation errors on (a) spacing and (b) speed for the seven models.**

To illustrate that the proposed DDPGvRT model can reproduce human-like driver behavior in dynamic situations such as approaching a standing vehicle, as well as stable car-following situations, two car-following periods outside the training dataset, one stable and the other dynamic, were randomly selected from the empirical data. Figure 8 shows the position, spacing gap, and speed observed in the empirical data, and the position, spacing gap, and speed replicated by the DDPGvRT, IDM, and RNN. It can be seen that the DDPGvRT model (in red) tracked the empirical (blue) gap and speed variations in both situations, and tracked them more closely than the IDM and RNN model did.

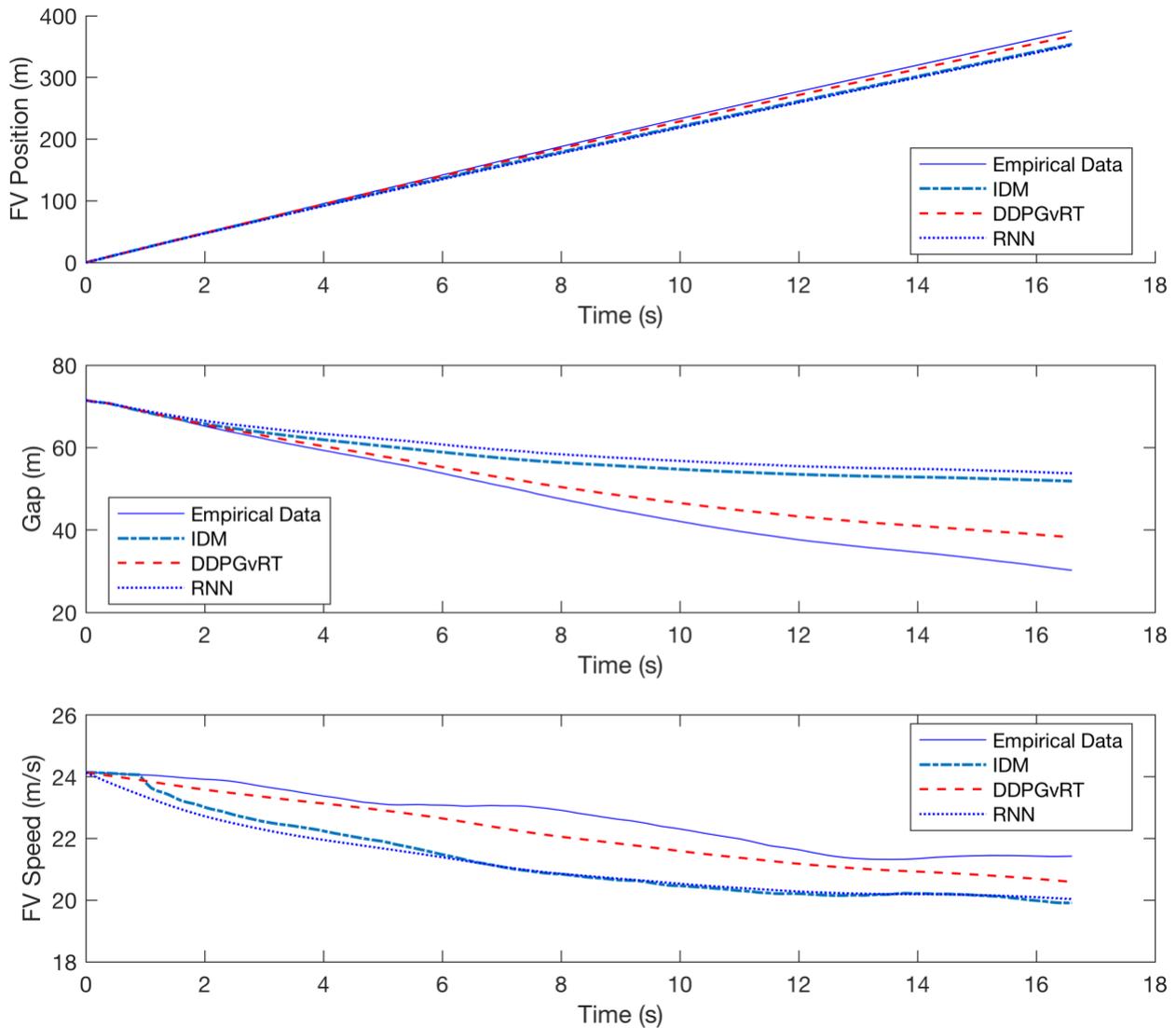

**(a) Stable Following**

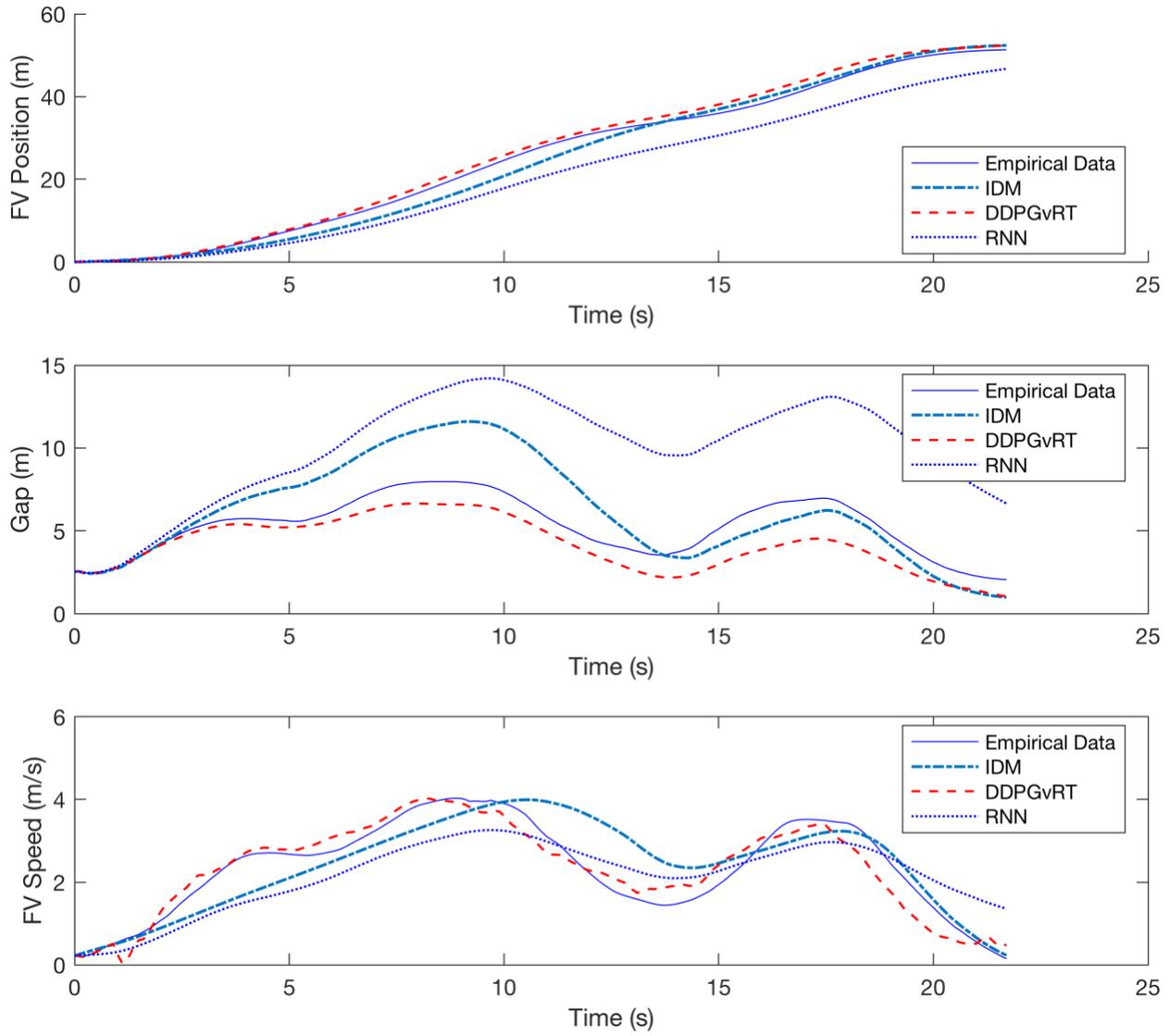

**(b) Dynamic Following**

**FIGURE 8 Car-following trajectories generated by the IDM, RNN, and DDPGv model.**

## 6.2 Generalization Capability

In the inter-driver validation phase, the car-following models were first calibrated for an investigated driver and were then applied to each of the remaining 19 drivers. Therefore, two 20 ×20 error matrices, one for spacing and the other for speed, were obtained for each model, as shown in Figure 9. The low RMSPE of the DDPGvRT model in both matrices indicates the DDPGvRT demonstrated the best generalization capability across different drivers.

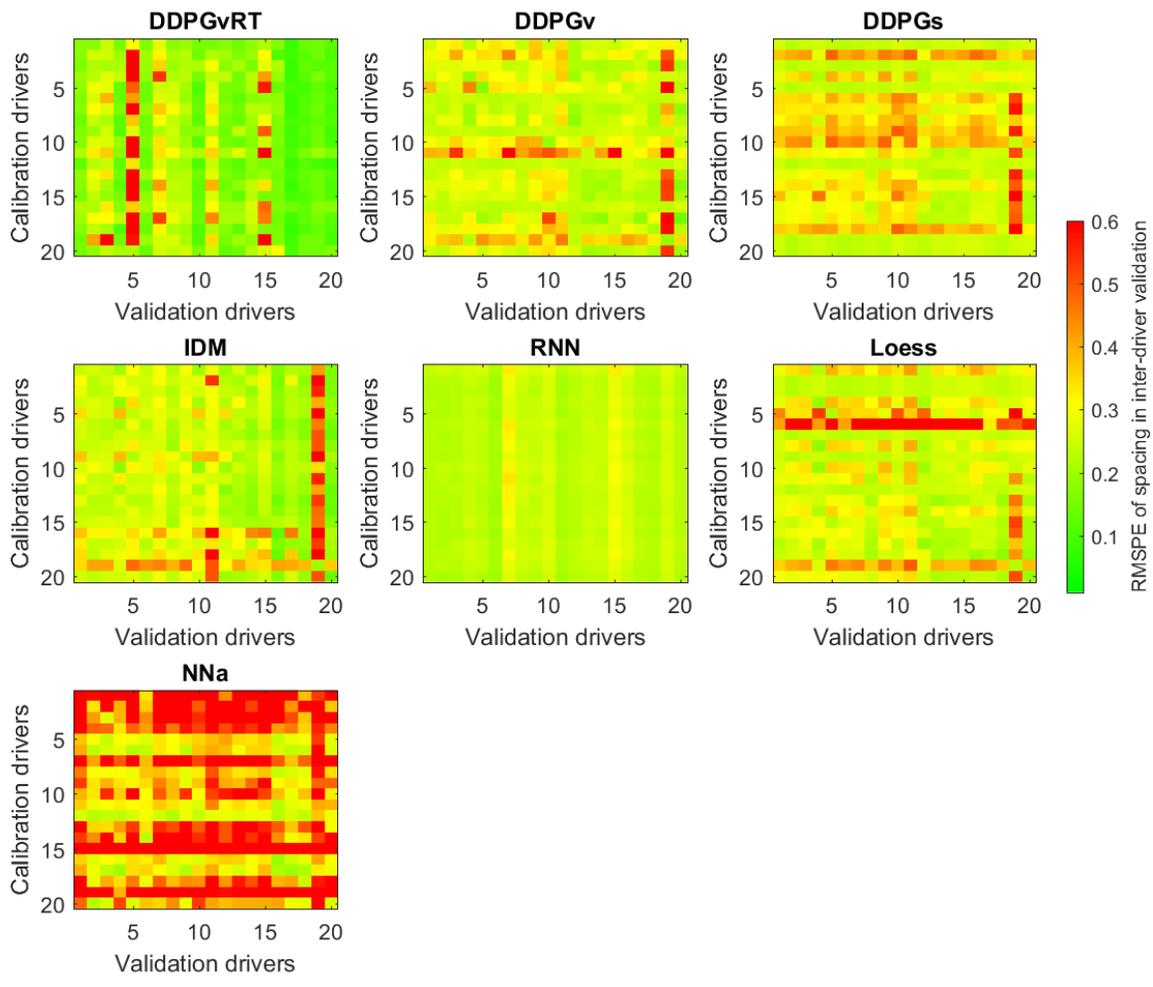

**(a) Spacing**

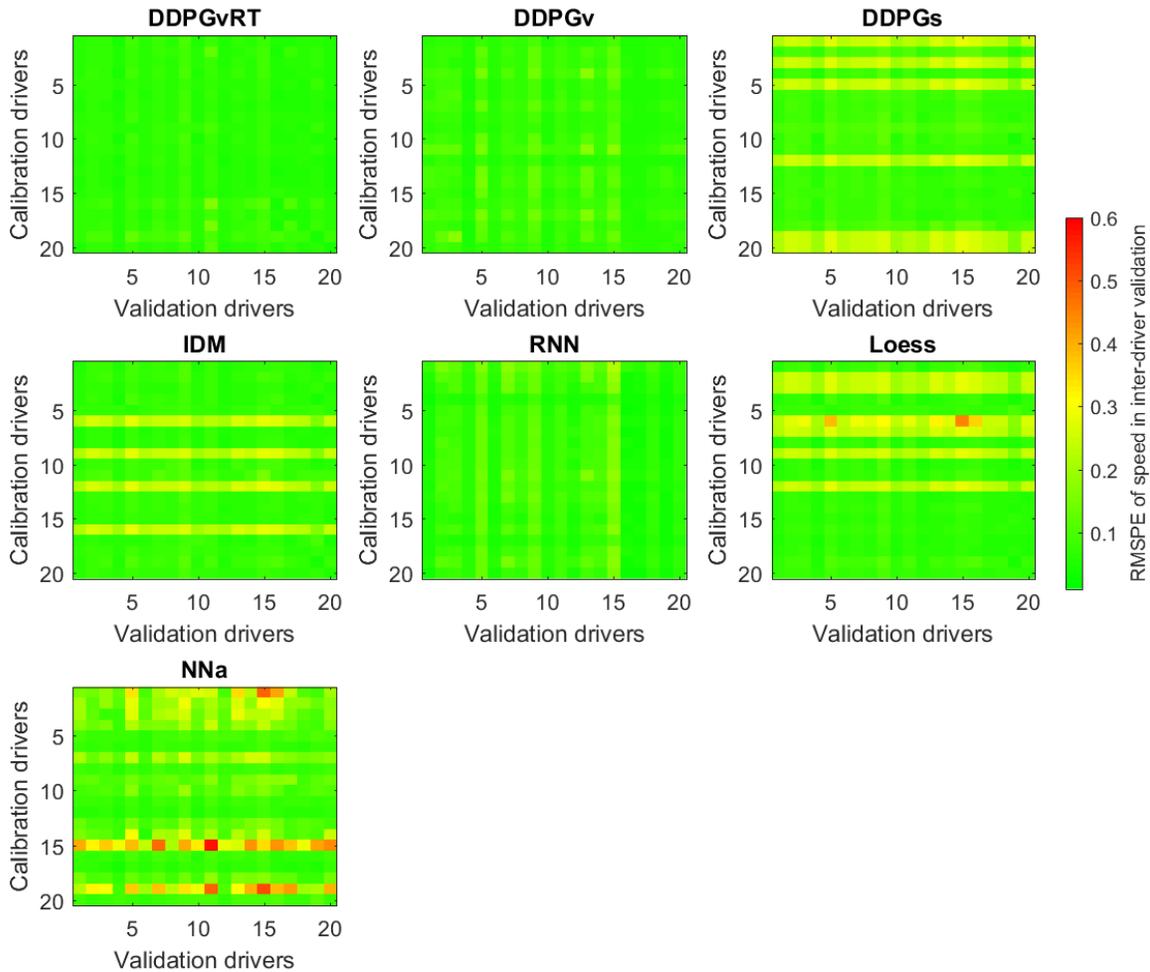

**(b) Speed**

**FIGURE 9 Inter-driver validation errors of (a) spacing and (b) speed for all pairs of drivers.**

Referring back to Figure 7 in Section 6.1, which illustrates the inter-driver validation errors for the seven models as well as the intra-driver errors discussed in that section, the DDPGvRT model showing the lowest mean values and standard deviation errors for both spacing and speed again makes clear that the DDPGvRT model outperformed all the investigated car-following models.

To identify whether low error values are expected for inter-driver validation across drivers with similar characteristics, the DDPGvRT model's mean values of inter-driver validation spacing errors across different types of drivers are presented in Table 4. As can been seen, intra-group validation errors are lower than those of inter-group validation, indicating that models calibrated by the data of a certain driver are more easily generalized to drivers who have similar driving styles to that calibration driver.

**TABLE 4 Mean Inter-Driver Validation Errors on Spacing across Different Types of Drivers for the DDPGvRT Model**

| Calibration Driver Type | Validation Driver Type | |
|---|---|---|
| | Aggressive | Conservative |
| Aggressive | 0.1984 | 0.2398 |
| Conservative | 0.2705 | 0.2259 |

## 6.3 Adaptivity

This section investigates how a pre-trained DDPG model can adapt to a new driver when it is retrained with that driver's data. Three drivers from the investigated 20 drivers were randomly selected, and three corresponding DDPGvRT models were trained with their individual driving data. Each model, corresponding to one of the three drivers, was then recalibrated with driving data from two other drivers, and, for completeness, with the corresponding calibrated driver. The retraining proceeded as follows for each model: 1) the old training cases in the replay buffer were cleared; 2) the agent was then fed with data from the new driver, and model training was implemented as described in Algorithm in Section 4.6.

Figure 10 presents the changes in spacing and speed errors during the retraining process. The vertical axis refers to corresponding drivers, on whose data the model was initially trained, and the horizontal axis refers to target drivers on whose data the model was retrained. As shown in the figure, the error curves begin with a stable horizontal line, a result of the RL agent not starting learning (updating network parameters) until the replay buffer was filled with new experience data. Then, a sharp rise in error succeeding the stable horizontal line can be observed in most episodes, a rise caused by exploration noise. Eventually, the errors declined and started to converge after approximately 8 episodes of training. The final errors were smaller than those in the initial models, which demonstrates that the DDPGvRT model can adapt to different drivers when new data are fed.

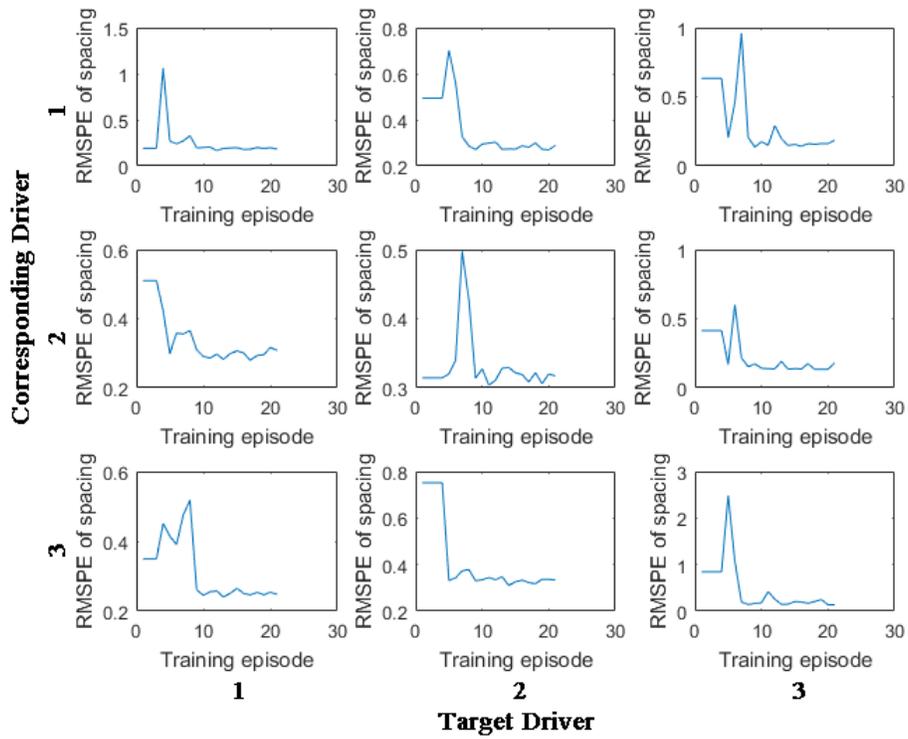

**(a) Spacing**

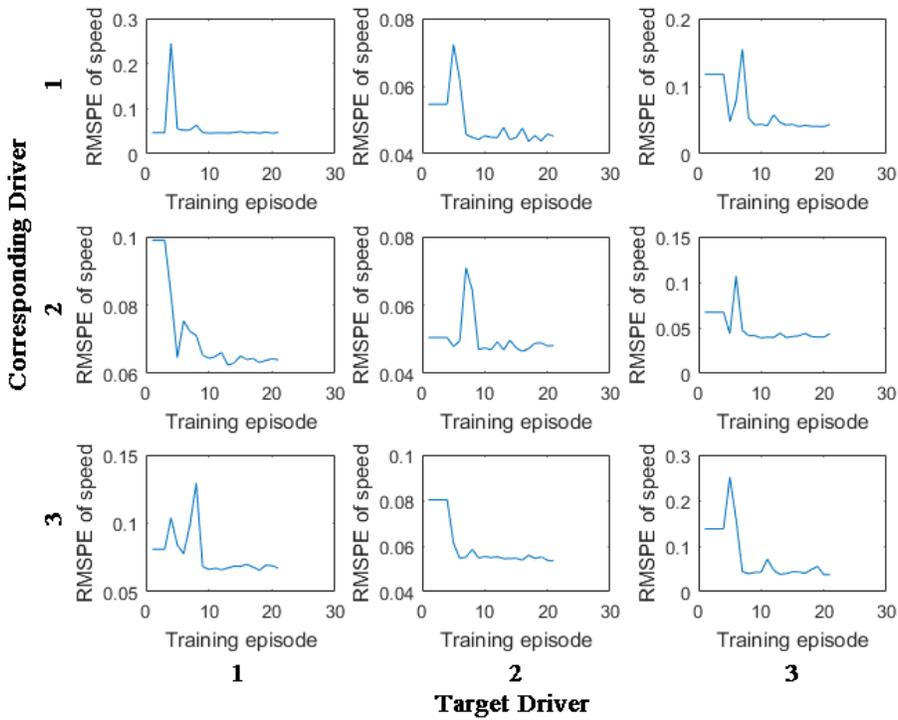

**(b) Speed**

**FIGURE 10 Changes in (a) spacing and (b) speed errors during retraining.**

# 7   DISCUSSION AND CONCLUSION

This study proposes a framework for human-like autonomous car-following planning based on deep reinforcement learning (deep RL). The framework uses a deep deterministic policy gradient (DDPG) algorithm to learn three types of car-following models, DDPGs, DDPGv, and DDPGvRT, from historical driving data. Results show that the proposed DDPGvRT model, which uses speed (velocity) deviation as the reward function and considers a reaction time delay of 1s, outperforms the tested traditional and recent data-driven car-following models in terms of trajectory-reproducing accuracy and generalization capability, and demonstrates its ability to adapt to different drivers.

The better performance of the DDPGvRT model in comparison to nonparametric models and conventional neural network-based models demonstrates that reinforcement learning can achieve better generalization capability because the agent learns decision-making mechanisms from training data rather than parameter estimation through fitting the data.

Three reasons may account for the better performance of the DDPGvRT in comparison to traditional car-following models:

1) Rule-based traditional car-following models have strong built-in specific assumptions about road conditions and driver behavior, which may not be consistent with reality. In contrast, the DDPG model's assumptions are more general: the model assumes only that drivers select actions to maximize cumulative future rewards, which agrees with psychological and neuroscientific perspectives of human behavior (*11*).

2) The small set of parameters used in parametric traditional models may not generalize well to diverse driving scenarios, while the DDPG model uses two neural networks, the actor for policy generation and the critic for policy improvement. The significantly greater number of parameters in neural networks improves the generalization capability of the DDPG model.

3) By adopting reaction time and value function (the expected reward for choosing a certain driving maneuver in current driving condition), the DDPGvRT model considers not only the instantaneous traffic conditions used in traditional car-following models, but also historical driving behaviors and long-term future rewards of current driving strategies.

An important finding (illustrated in Figure 7) is that inter-driver validation errors consistently exceed intra-driver errors, which indicates that considerable behavioral differences exist between different drivers, as previously discussed by Ossen and Hoogendoorn (*39*). This confirmation justifies the need for autonomous driving algorithms to adapt to different drivers by continuously learning. Furthermore, the difference in validation errors suggests that inter-driver heterogeneity in car-following should not be neglected in microscopic traffic simulation, but that there is a need to use various driver archetypes, such as aggressive and conservative, to build a traffic mix.

Although the network structure utilized in our study with only one hidden layer cannot be regarded as a truly "deep" neural network, it fully exploits the main idea of deep reinforcement learning methods and can be easily extended once more input variables are provided. A potential improvement to our model is related to experience replay, which enables RL agents to learn from past experiences. In the currently adopted DDPG algorithm, experience transitions were sampled from a replay memory at the same frequency that they were originally experienced, with the consequence that their potential significance was ignored by the RL agent (*43*). A framework for prioritizing experience can be used in future work to facilitate more efficient learning by replaying important transitions more frequently.

Moreover, since the DDPG algorithm can learn policies based on low-dimension inputs and also directly from raw pixel inputs (end-to-end fashion) (*36*), we would expect to compare the performance of these two paradigms in future studies. By comparing the low-dimension input-based DDPG car-following model, which accepts relative position and speed information, with an image-based end-to-end DDPG model, we may reveal the latent relationship between human perception and car-following controlling action.

As for the training time issue, first, the model only needs a few seconds to converge for data collected on trajectories lasting many seconds. Second, when an initial optimal policy has been learned, that policy is responsible for real-time referenced acceleration generation, which is feasible because it is just a forward pass on the policy network; the policy updating work, however, does not necessarily need to be conducted in real time.

It is worth noting that the proposed DDPG model is intended to reproduce human-like driving behavior in order to facilitate more uniform and therefore more predictable interaction of autonomous vehicles with human-driven vehicles; absolute safety issues, as well as the handling of human error in historical driving data, are not addressed in this approach. Safety issues must be addressed within the province of a coupled collision avoidance system, and human errors can then be filtered by an anomaly detection algorithm. Future research will refine the DDPG approach and validate it against field experiments. The results of this study, however, give an indication of what is feasible.

## ACKNOWLEDGEMENTS

This study was jointly sponsored by the Chinese National Science Foundation (51522810; 51878498), the Science and Technology Commission of Shanghai Municipality (18DZ1200200), and the 111 Project (B17032).

## APPENDIX: DESCRIPTION OF THE COMPARISON MODELS

### A. Loess car-following model

Papathanasopoulou and Antoniou (*28*) developed a nonparametric data-driven car-following model based on locally weighted regression, the Loess model, as shown in Equation (*A.1*):

$$y_i = g(x_i) + \varepsilon_i \qquad\qquad (A.1)$$

where $y_i$ is the response variable (e.g., following vehicle's acceleration); $x_i$ are predictor variables (vectors containing relative speed, gap, and following vehicle's speed); $i = 1, 2, …, $ n-1, where n is the index of observations; $g$ is the regression function (a linear function was used in Papathanasopoulou and Antoniou (*28*)); and $\varepsilon_i$ are residual errors.

Local regression relies on the idea that near $x=x_0$ the response variable could be approximately estimated by fitting a regression surface to the data points located within the neighborhood of the point $x_0$, which is restricted by a smoothing parameter: span $\alpha$. The span defines the smoothness of the estimated surface as it specifies the percentage of data that would be taken into consideration for each local fit (span = 0.4 in Papathanasopoulou and Antoniou (*28*)). The data included in the "area of influence" are weighted according to their Euclidean distance from the center of neighborhood $x$. In Papathanasopoulou and Antoniou (*28*), a tri-cube weight function was used:

$$W(u) = \begin{cases} (1-u^3)^3, & 0 \le u \le 1 \\ 0, & otherwise \end{cases} \tag{A.2}$$

where $u$ is a variable that will be used to estimate weights.

Then, the weight of each observation ($y_i$, $x_i$) is:

$$w_i(x) = W(\frac{p(x, x_i)}{dist(x)}) \tag{A.3}$$

where $x_i$ is the vector of the observation variables, $x$ is the center of the chosen neighborhood, $p(x, x_i)$ is the Euclidian distance between $x$ and $x_i$, and $dist(x)$ is the distance of the most distant predictor value $x_i$ to $x$ within the chosen neighborhood.

Taking into account the calculated weights, the Loess method is defined through an optimization problem. The objective function that should be minimized is the weighted residual sums of squares:

$$\sum_{i=1}^{n} w_i \cdot \varepsilon_i^2 = \sum_{i=1}^{n} w_i \cdot (y_i - x_i \cdot \beta)^2 \tag{A.4}$$

where $y_i$ is the response variable and $\beta$ is the parameter of the polynomial. At each $x$, the value of the parameter $\beta$ that minimizes Equation (*A.4*) is found.

### B. Conventional neural network car-following model

A conventional feedforward neural network that predicts followers' acceleration based on car-following gap and speed was implemented (referred to as the NNa model). The implemented NNa model had a network architecture the same as that of DDPGv's actor network, as shown in the following figure:

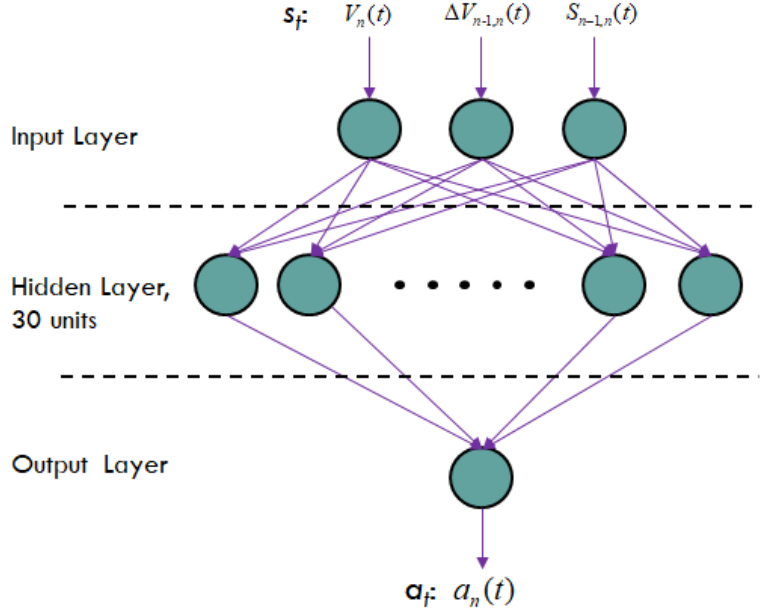

**FIGURE A.1 Network architecture for the NNa model.**

At time step *t*, the actor network takes a state $s_t = (v_n(t), \Delta v_{n-1,n}(t), \Delta S_{n-1,n}(t))$ as input, and outputs a continuous action: the following vehicle's acceleration $a_n(t)$. Once acceleration is calculated, car-following states (following gap, speed, and relative speed) are updated according to Equation (*3*). The neural network was trained by minimizing the differences between observed acceleration values and those predicted by the network.

*C. Recurrent neural network (RNN) based car-following model*

A recurrent neural network (RNN) based car-following model proposed by Zhou et al. (*32*) was implemented as a comparison model. The RNN has an internal state that represents the current situation. RNNs can build their internal memory as the foundation for the next prediction so as to process sequences of inputs.

A typical RNN's architecture is shown in Figure A.2. It is clear that RNNs take a sequence of inputs to generate another sequence of outputs. All inputs and outputs are arranged in order. Therefore, RNNs learn the hidden sequence order as well as the corresponding output value.

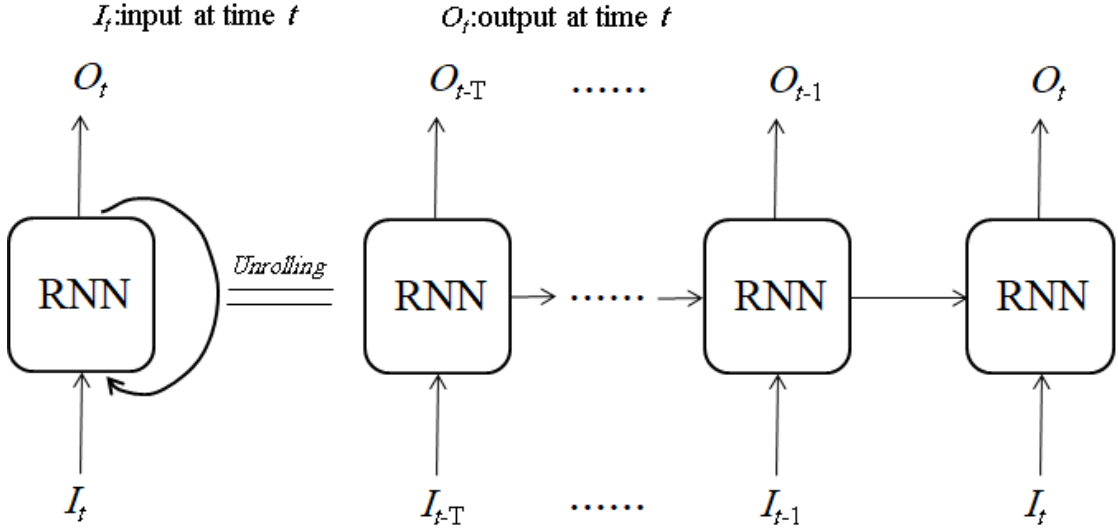

**FIGURE A.2 A typical RNN architecture, folded RNN (left) and unfolded RNN (right) (Zhou et al.**
**(*32*)).**

For RNNs, the equations for computing the output are also upgraded as follows:

$$h_t = \text{ReLU}(W_h h_{t-1} + W_i I_t + b_i) \tag{C.1}$$

$$O_t = W_o h_t + b_o \tag{C.2}$$

where $W_i$ and $b_i$ denote the input weights and biases, $h_t$ is the hidden state that represents the internal memory at time $t$, $I_t$ represents the input to the network at time $t$, $W_h$ are the weights of the hidden state, $W_o$ and $b_o$ stand for the output weights and biases, and $O_t$ represents the output at time $t$. The ReLu is a nonlinear activation function as discussed in Section 4.3.

In Zhou et al. (*32*), the RNN based car-following model takes inputs including gap, relevant speed and vehicle speed at time step $t$ and output follower acceleration for the next time step. Then, the follower's positions and velocity for the next step are updated based on Equation (*3*). The number of hidden units is set as 60.

The objective function for the RNN model is defined as:

$$C(W,b) = \frac{\left(S_{n-1,n}(t) - S_{n-1,n}^{obs}(t)\right)^2}{\left(S_{n-1,n}^{obs}(t)\right)^2} \tag{C.3}$$

where $S_{n-1,n}(t)$ is the simulated spacing and speed at time step $t$, and $S_t^{obs}$ is the observed spacing in the empirical data set at time step $t$; and $W$ and $b$ denote current weights and biases in the RNN model. The RNN model gradually minimizes this objective function by back-propagating a small update through time in the direction of optimizing the weights and biases.